\documentclass[sigconf]{acmart}
\copyrightyear{2022}
\acmYear{2022}
\setcopyright{acmcopyright}

\acmConference[CIKM '22] {Proceedings of the 31st ACM International Conference on Information and Knowledge Management}{October 17--21, 2022}{Atlanta, GA, USA.}
\acmBooktitle{Proceedings of the 31st ACM International Conference on Information and Knowledge Management (CIKM '22), October 17--21, 2022, Atlanta, GA, USA}
\acmPrice{15.00}
\acmDOI{10.1145/3511808.3557647}
\acmISBN{978-1-4503-9236-5/22/10}

\usepackage{hyperref}
\usepackage{url}

\usepackage{multirow}
\usepackage{graphicx}
\usepackage{amsmath}
\usepackage{amsthm}
\usepackage{mathtools}

\usepackage{lipsum}
\usepackage{enumitem}

\usepackage{algorithm2e}
\RestyleAlgo{ruled}
\SetKwComment{Comment}{\# }{}

\usepackage{caption}
\usepackage{subcaption}
\usepackage{CJKutf8}
\usepackage{soul}

\usepackage{wrapfig}
\usepackage{xcolor}
\usepackage{colortbl}
\usepackage{tabularx}
\usepackage{stfloats}

\usepackage{balance} 

\usepackage{amsmath,amsfonts,bm}

\def\eqref#1{equation~\ref{#1}}

\def\1{\bm{1}}

\def\vd{{\vec{d}}}

\def\vg{{\vec{g}}}
\def\vh{{\vec{h}}}

\def\vs{{\vec{s}}}

\def\vy{{\vec{y}}}

\def\mH{{\bm{H}}}

\def\mX{{\bm{X}}}

\DeclareMathAlphabet{\mathsfit}{\encodingdefault}{\sfdefault}{m}{sl}
\SetMathAlphabet{\mathsfit}{bold}{\encodingdefault}{\sfdefault}{bx}{n}

\def\gD{{\mathcal{D}}}
\def\gE{{\mathcal{E}}}
\def\gF{{\mathcal{F}}}
\def\gG{{\mathcal{G}}}

\def\gI{{\mathcal{I}}}

\def\gL{{\mathcal{L}}}

\def\gN{{\mathcal{N}}}

\def\gR{{\mathcal{R}}}
\def\gS{{\mathcal{S}}}

\def\sA{{\mathbb{A}}}
\def\sB{{\mathbb{B}}}

\def\sR{{\mathbb{R}}}
\def\sS{{\mathbb{S}}}

\def\sV{{\mathbb{V}}}

\def\sY{{\mathbb{Y}}}

\newcommand{\E}{\mathbb{E}}

\newcommand{\softmax}{\mathrm{softmax}}
\newcommand{\sigmoid}{\sigma}

\newcommand*{\scale}[2][4]{\scalebox{#1}{$#2$}}%

\newcommand{\cmt}[1]{} 

\newcommand{\pvalue}[0]{\textit{p}-value}
\newcommand{\ttest}[0]{\textit{t}-test}

\newcommand{\MLP}[0]{\mathrm{MLP}}

\newcommand{\idx}[0]{\texttt{idx}}
\newcommand{\topk}[0]{\mathrm{top}_k}

\newcommand{\textor}[0]{\textrm{or}}

\newcommand{\khop}[0]{k\text{-hop}}

\newcommand{\glob}[0]{\textrm{glob}}
\newcommand{\sub}[0]{\textrm{sub}}
\newcommand{\obs}[0]{\textrm{obs}}

\newcommand{\lambdakhop}[0]{\lambda^{\khop}}

\newcommand{\FNTN}[0]{\textsf{\fontsize{8.5pt}{8.5pt}\selectfont FNTN}}
\newcommand{\HPOMetab}[0]{\textsf{\fontsize{8.5pt}{8.5pt}\selectfont HPO-Metab}}
\newcommand{\EMUser}[0]{\textsf{\fontsize{8.5pt}{8.5pt}\selectfont EM-User}}
\newcommand{\FNTNb}[0]{\textsf{\fontsize{8.5pt}{8.5pt}\selectfont FNTN} }
\newcommand{\HPOMetabb}[0]{\textsf{\fontsize{8.5pt}{8.5pt}\selectfont HPO-Metab} }
\newcommand{\EMUserb}[0]{\textsf{\fontsize{8.5pt}{8.5pt}\selectfont EM-User} }

\newtheorem{proposition}{Proposition}

\definecolor{weakgray}{HTML}{e7e7e7}
\definecolor{weakblue}{HTML}{E3F2FD}
\definecolor{weakred}{HTML}{FFEBEE}
\definecolor{midblue}{HTML}{bbdefb}
\definecolor{midred}{HTML}{ffcdd2}

\newcommand\cellwg{\cellcolor{weakgray}}
\newcommand\cellwb{\cellcolor{weakblue}}
\newcommand\cellwr{\cellcolor{weakred}}
\newcommand\cellmb{\cellcolor{midblue}}

\definecolor{obspink}{HTML}{EF5FA7}
\definecolor{subblue}{HTML}{00A2FF}

\newcommand{\propositioncondgan}[0]{
\begin{proposition}[The conditional GAN-like divergence MI bound]\label{prop:conditional-gan-mi}
For $d$-dimensional random variables $X$ and $Y$ with a joint distribution $p(x, y)$ and marginal distributions $p(x)$ and $p(y)$, fix any function $f: (X, Y) \rightarrow \sR$ and realization $x$ of $X$. Let $c_x = \E_{y \sim p(y)} \left[ e^{f(x, y)} \right]$, $\sB_{c_x} \subset \sR$ be strictly lower bounded by $c_x$, and $\sY_{c_x} = \{ y \vert e^{f(x, y)} \in \sB_{c_x} \}$ with an assumption of $p(\sY_{c_x}) > 0$. For $\sY_{r}$ in the Borel $\sigma$-algebra over $\sR^d$, let $q( Y \in \sY_{r} \vert X = x ) = p(\sY_{r} \vert \sY_{c_x})$, then $\gI^{\textrm{CGD}} \leq \gI^{\textrm{GD}}$ where
\begin{align}
& \qquad\quad \begin{aligned}
    \mathllap{\gI^{\textrm{CGD}}} =\ & \E_{x, y \sim p(x, y)} \left[ \log \sigmoid ( f (x, y)) \right] + \\
    & \E_{x \sim p(x)} \E_{y \sim q(y \vert x)} \left[ \log \left( 1 - \sigmoid ( f (x, y)) \right) \right],
\end{aligned} \\
& \qquad\quad \begin{aligned}
    \mathllap{\gI^{\textrm{GD}}} =\ & \E_{x, y \sim p(x, y)} \left[ \log \sigmoid ( f (x, y)) \right] + \\
    & \E_{x \sim p(x)} \E_{y \sim p(y)} \left[ \log \left( 1 - \sigmoid ( f (x, y)) \right) \right].
\end{aligned}
\end{align}
\end{proposition}
}

\begin{document}

\title{Models and Benchmarks for Representation Learning of Partially Observed Subgraphs}


\author{Dongkwan Kim}
\email{dongkwan.kim@kaist.ac.kr}
\affiliation{%
  \institution{Korea Advanced Institute of Science and Technology}
  \city{Daejeon}
  \country{Republic of Korea}
}

\author{Jiho Jin}
\email{jinjh0123@kaist.ac.kr}
\affiliation{%
  \institution{Korea Advanced Institute of Science and Technology}
  \city{Daejeon}
  \country{Republic of Korea}
}

\author{Jaimeen Ahn}
\email{jaimeen01@kaist.ac.kr}
\affiliation{%
  \institution{Korea Advanced Institute of Science and Technology}
  \city{Daejeon}
  \country{Republic of Korea}
}

\author{Alice Oh}
\email{alice.oh@kaist.edu}
\affiliation{%
  \institution{Korea Advanced Institute of Science and Technology}
  \city{Daejeon}
  \country{Republic of Korea}
}

\renewcommand{\shortauthors}{Dongkwan Kim, Jiho Jin, Jaimeen Ahn, \& Alice Oh}




\begin{abstract}
Subgraphs are rich substructures in graphs, and their nodes and edges can be partially observed in real-world tasks. Under partial observation, existing node- or subgraph-level message-passing produces suboptimal representations. In this paper, we formulate a novel task of learning representations of partially observed subgraphs. To solve this problem, we propose Partial Subgraph InfoMax (PSI) framework and generalize existing InfoMax models, including DGI, InfoGraph, MVGRL, and GraphCL, into our framework. These models maximize the mutual information between the partial subgraph's summary and various substructures from nodes to full subgraphs. In addition, we suggest a novel two-stage model with $k$-hop PSI, which reconstructs the representation of the full subgraph and improves its expressiveness from different local-global structures. Under training and evaluation protocols designed for this problem, we conduct experiments on three real-world datasets and demonstrate that PSI models outperform baselines.
\end{abstract}
\maketitle

\section{Introduction}\label{sec:introduction}
The graph neural network (GNN) has become a major framework to learn representations of nodes, edges, and graphs~\citep{bronstein2017geometric, battaglia2018relational, hu2020open, dwivedi2020benchmarking}. In addition, subgraphs can express various real-world data: news propagation in a social network or disease in a graph of symptoms~\citep{alsentzer2020subgraph}.

The current formulation of subgraph representation learning by \citet{alsentzer2020subgraph} assumes full observation of nodes and edges in a subgraph, and that assumption often does not hold in the real world. If so, existing models may learn suboptimal representations because of inaccurate message-passing with missing nodes or edges. In this paper, we suggest a novel task of learning representations of partial subgraphs by relaxing the assumption of complete observation.

For this `partial subgraph learning' task, we propose the \textit{Partial Subgraph InfoMax} (PSI) framework based on mutual information (MI) maximization. Inspired by Deep InfoMax~\citep{hjelm2018learning} that maximizes MI between the global summary (e.g., images) and local parts (e.g., patches), PSI maximizes MI between a partial subgraph and its substructures (e.g., nodes or full subgraphs). We generalize existing InfoMax models for node and graph-level tasks~\citep{velickovic2018deep, Sun2020InfoGraph:, hassani2020contrastive, you2020graph} to solve the partial subgraph learning problem. PSI models first summarize a specific partial subgraph and learn to distinguish for its summary whether a substructure is related to the same subgraph (positive) or not (negative). This allows learning the structural hierarchy of nodes, partial and full subgraphs in subgraph representations.

However, the summary of the partial subgraph cannot explicitly encode missing information. Thus, we employ two-stage PSI models that reconstruct the summary close to the full subgraph under insufficient observation. We first propose `$k$-hop PSI' that reconstructs the full subgraph by assembling $k$-hop neighbors of high MI with the partial subgraph. Then, the second PSI model takes the reconstructed subgraph summary as input and learns local-global structures different from the first $k$-hop PSI.

\begin{figure}[t]
  \centering
  \vspace{-0.1cm}
  \includegraphics[width=\columnwidth]{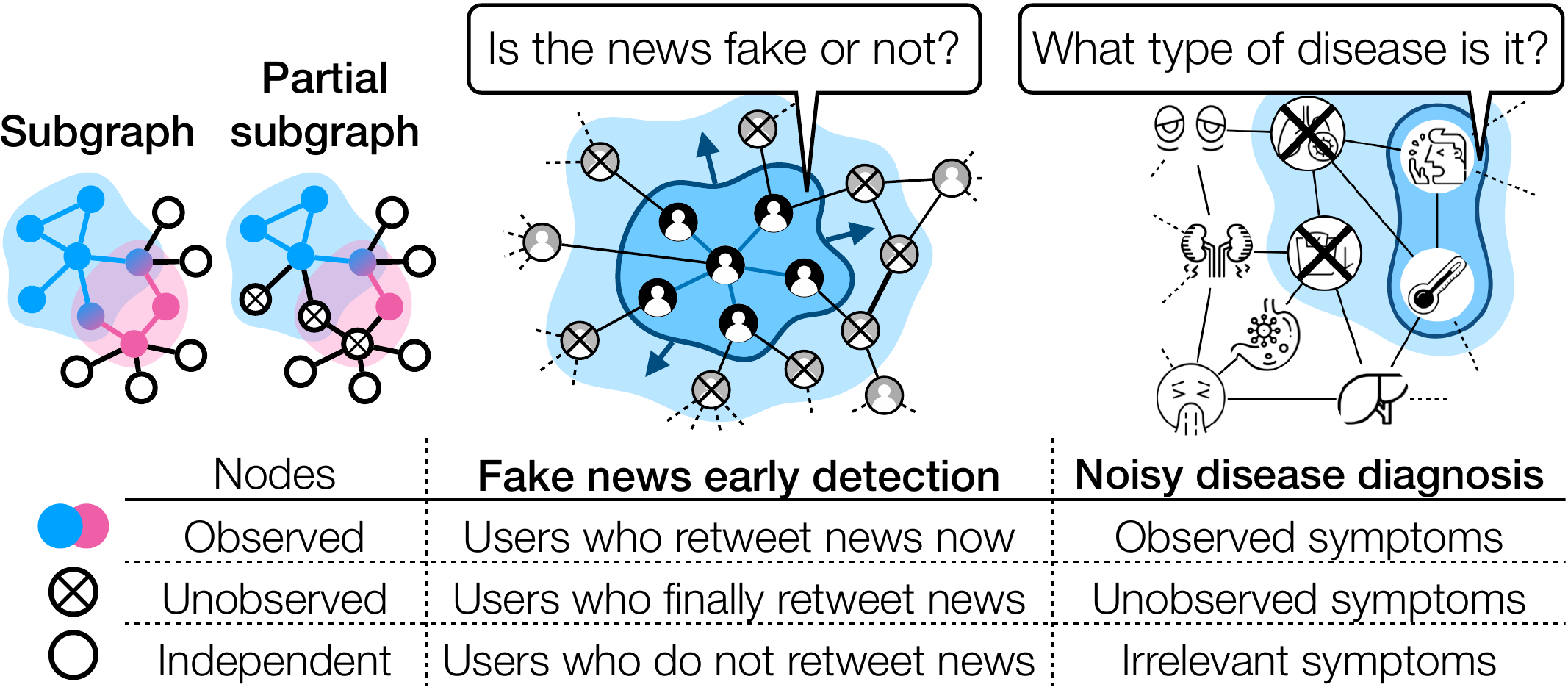}
  \vspace{-0.7cm}
  \caption{
  Left: Comparison of subgraph and partial subgraph representation learning. Middle \& Right: Real-world examples (\S\ref{paragraph:real-world-examples}) of partial subgraph learning.
  }
  \label{fig:problem}
  \vspace{-0.6cm}
\end{figure}

We demonstrate the improved representation learning performance of PSI models with experiments on three real-world datasets. These datasets simulate scenarios of fake news early detection, social network user profiling, and disease diagnosis with partial observation (Figure~\ref{fig:problem}). Our models consistently outperform baseline models for all datasets. In addition, we analyze models' performance depending on the properties of subgraphs and the global graph.

We present the following contributions. First, we formulate the partial subgraph learning problem and suggest realistic training and evaluation protocols (\S\ref{sec:problem}). Second, we propose the Partial Subgraph InfoMax framework for this problem (\S\ref{sec:model}). Third, we demonstrate that our model outperforms baselines on three real-world datasets (\S\ref{sec:results}). Our code is available at \url{https://github.com/dongkwan-kim/PSI}.


\vspace{-0.22cm}
\section{\fontsize{10.8}{15}\selectfont Partial Subgraph Learning Problem}\label{sec:problem} 
\vspace{-0.1cm}
We formulate a novel problem of learning subgraph representations under partial observations of nodes and edges. 

\vspace{-0.22cm}
\paragraph{Notations}

Let $\gG = (\sV^{\glob}, \sA^{\glob})$ be a global graph, where $\sV^{\glob}$ is a set of nodes and $\sA^{\glob}$ is a set of edges, and $\mX^{\glob} \in \sR^{|\sV^{\glob}| \times F^{\mathrm{in}}}$ is a feature matrix of nodes. A subgraph $\gS = (\sV^{\sub}, \sA^{\sub})$ of $\gG$ is defined as a graph, the nodes and edges of which are subsets of $\sV^{\glob}$ and $\sA^{\glob}$ respectively. Each subgraph has a label $y \in \{ 1, ..., C \}$, and sometimes, a subgraph-level feature $\vg \in \sR^{F'}$ may exist.

\vspace{-0.22cm}
\paragraph{Problem formulation} \label{sec:problem_formulation}

We formulate the `partial subgraph learning' by relaxing the complete observation assumption of \citet{alsentzer2020subgraph}, considering a subset of nodes or edges of the subgraph, as in Figure~\ref{fig:problem}.
We define a partial subgraph $\gS^{\obs}$ of $\gS$ as $\gS^{\obs} = (\sV^{\obs}, \sA^{\obs})$ where $\sV^{\obs} \subset \sV^{\sub}$ and $ \sA^{\obs} \subset \sA^{\sub}$.
We denote a set of subgraphs as $\sS = \{ \gS_1, ..., \gS_M \}$, and corresponding partial subgraphs as $\sS^{\obs} = \{ \gS^{\obs}_1, ..., \gS^{\obs}_M \}$, where each $\gS^{\obs}_i$ is a subgraph of $\gS_i$ for $i \in \{1, ..., M \}$.
We aim to learn a representation $\vs \in \sR^F$ for each $\gS^{\obs} \in \sS^{\obs}$ to predict $y$.

\vspace{-0.22cm}
\paragraph{Real-world examples}\label{paragraph:real-world-examples}

One example of a real-world scenario is in detection of fake news in a social network, where the propagation tree of news can be represented as a subgraph. 
Rather than a fully propagated subgraph, it is more useful to detect fake news with an early propagated subgraph before the news spreads.
Here, nodes are observed according to the order in which the information is propagated, and a partial subgraph would contain only nodes of the early propagation. Another example is a noisy diagnosis task of diseases (as subgraphs) based on a knowledge graph of symptoms (as nodes)~\citep{alsentzer2020subgraph}. In some cases, not all symptoms of the disease may appear or be seen, and a diagnosis is made based solely on the observed ones.
The observation order of symptoms would not follow a fixed order but depend on the specific situation of the patient.


\vspace{-0.22cm}
\paragraph{Realistic training and evaluation protocols} 

For training and evaluation, we create partial subgraphs by selecting nodes $\sV^{\obs}$ from the nodes $\sV^{\sub}$ of the full subgraph. 
We fix validation and test node sets with a \textit{constant} size for each subgraph. 
This is more realistic than selecting nodes in proportion to the subgraph size (i.e., $|\sV^{\sub}|$) in that we cannot know the exact size at evaluation. 
We create a new partial subgraph of fixed size for training at every iteration.
As we note above, there may be a specific observation ordering of the nodes.
It is natural to take these into account when constructing the partial subgraphs. 
Thus, we select early observed nodes if the observation is \textit{ordered}, otherwise we sample the nodes randomly.

\vspace{-0.22cm}
\section{Models}\label{sec:model}
\vspace{-0.1cm}
We introduce the Partial Subgraph InfoMax (PSI) framework and its models. 
We first describe encoder-readout pipelines for learning subgraphs. 
Given a subgraph $(\sV^{*}, \sA^{*})$ and features $\mX^{\glob}$, the GNN encoder $\gE$ outputs the node representations $\mH^{*} = [\, \vh_1 \,|\, ... \,|\, \vh_{|\sV^{*}|} ]^\top \in \sR^{|\sV^{*}| \times F}$, and the readout $\gR$ generates the summary $\vs^{*} \in \sR^F$. Finally, the prediction function $\gF$ computes the logit $\vy \in \sR^C$. The superscript $*$ denotes the graph type as in \S\ref{sec:problem_formulation} such as `sub' for the full subgraph, `obs' for the partial subgraph. This is summarized as:
\vspace{-0.35cm}
\begin{align}
    (\sV^{*}, \sA^{*}, \mX^{\glob}) \xmapsto{\gE} \mH^{*} \xmapsto{\gR} \vs^{\,*} \xmapsto{\gF} \vy.
\end{align}

\vspace{-0.25cm}
\subsection{Partial Subgraph InfoMax framework}
\vspace{-0.1cm}

The encoder-readout is insufficient for partial subgraph learning since only information from observed nodes is considered for prediction.
Thus, we leverage a structural hierarchy from nodes to partial and full subgraphs using mutual information (MI) maximization.
The PSI framework encodes the information of the full subgraph into the partial subgraph representation by maximizing MI between the partial subgraph summary $\vs^{\obs}$ and representations from the full subgraph, nodes $\mH^{\sub}$ or summary $\vs^{\sub}$.

Among several MI estimators modeled with neural networks~\citep{belghazi2018mutual}, GAN-like divergence (GD)~\citep{nowozin2016fgan} and InfoNCE~\citep{oord2018representation} estimators are widely used in InfoMax models for graphs. To maximize GAN-like divergence estimator between nodes $\mH^{\sub}$ and the subgraph $\vs^{\obs}$, we minimize the following loss between samples from the joint distribution $P$ and the product of marginal distributions $P \times \Tilde{P}$:
\vspace{-0.1cm}
\begin{align}\label{eq:infomax_loss}
\scale[0.86]{
\gL^{\textrm{GD}} =
- \E_{P} \left[
    \log \sigma \left(
        \gD ( \vh_s, \vs^{\obs} )
    \right)
\right] -
\E_{P \times \Tilde{P}} \left[
    \log  \left(
        1 - \sigma \left(
            \gD ( \vh_{\Tilde{s}}, \vs^{\obs} )
        \right)
    \right)
\right],
}
\end{align}
where $s$ is an input sample from an empirical distribution $P$ of the input space, $\Tilde{s}$ is a negative sample from $\Tilde{P}$, $\vs^{\obs}$ is the partial subgraph summary of $s$, $\vh_s$ is the node representation in $s$. 
A discriminator $\gD : \sR^{F} \times \sR^{F} \rightarrow \sR$ computes how much $\vh_s$ and $\vs$ are related. 
We also maximize the MI bound by minimizing InfoNCE loss,
\vspace{-0.1cm}
\begin{align}\label{eq:infonce_loss}
\scale[0.89]{
\gL^{\textrm{InfoNCE}} =
\E_{P} \left[
    \gD \left( \vh_s, \vs^{\obs} \right) -
\E_{\Tilde{P}} \left[
     \textstyle
     \log \sum_{\Tilde{s}} e ^ {\gD ( \vh_{\Tilde{s}}, \vs^{\obs} )}
 \right]
 \right].
}
\end{align}

We generalize the following InfoMax models for learning partial subgraph representations: DGI~\citep{velickovic2018deep}, InfoGraph~\citep{Sun2020InfoGraph:}, MVGRL~\citep{hassani2020contrastive}, and GraphCL~\citep{you2020graph} that maximize MI between local and global structures in the graph. Since they are designed for node or graph predictions, we incorporate them into PSI by considering each partial subgraph as an independent graph. Then, we jointly minimize the InfoMax loss and the cross-entropy loss $\gL^{\textrm{graph}}$ on the logit $\vy$ and label $y$. See Table ~\ref{tab:psi_models} for the architectures of PS-prefixed PSI models.

\begin{table}[]
\centering
\caption{The summary of differences between PSI models.}
\label{tab:psi_models}
\vspace{-0.375cm}
\resizebox{\columnwidth}{!}{%
\begin{tabular}{l|cccc}
\hline
                 & \multicolumn{1}{c|}{PS-DGI} & \multicolumn{1}{c|}{PS-MVGRL} & \multicolumn{1}{c|}{PS-InfoGraph} & PS-GraphCL \\ \hline
\begin{tabular}[c]{@{}l@{}}Substructures to\\ maximize MI with\\ partial subgraphs\end{tabular} &
  \multicolumn{3}{c|}{Nodes in full subgraphs} &
  \begin{tabular}[c]{@{}c@{}}Summary of\\ full subgraphs\end{tabular} \\ \hline
Negative samples & \multicolumn{2}{c|}{Row-wise shuffled nodes}                & \multicolumn{2}{c}{Nodes in other subgraphs}   \\ \hline
\begin{tabular}[c]{@{}l@{}}Graph\\ augmentations\end{tabular} &
  \multicolumn{1}{c|}{-} &
  \multicolumn{1}{c|}{\begin{tabular}[c]{@{}c@{}}Personalized\\ PageRank\end{tabular}} &
  \multicolumn{1}{c|}{-} &
  \begin{tabular}[c]{@{}c@{}}Node dropping,\\ edge perturbation,\\ attribute masking\end{tabular} \\ \hline
\begin{tabular}[c]{@{}l@{}}Shared encoders for\\ partial \& augmented\\ subgraphs\end{tabular} &
  \multicolumn{1}{c|}{N/A} &
  \multicolumn{1}{c|}{False} &
  \multicolumn{1}{c|}{N/A} &
  True \\ \hline
MI estimator     & \multicolumn{3}{c|}{GAN-like Divergence (GD)}                                                                         & InfoNCE    \\ \hline
\end{tabular}%
}
\vspace{-0.45cm}
\end{table}



\vspace{-0.25cm}
\subsection{Two-Stage Models with $k$-hop PSI}
\vspace{-0.1cm}

Summarizing only nodes in the partial subgraph does not explicitly include all feature and structure information in the full subgraph. 
We propose $k$-hop PSI that reconstructs the full subgraph based on the partial subgraph's $k$-hop neighborhoods using the MI estimator.

For subgraph reconstruction, $k$-hop PSI first inspects which nodes in $\gG$ belong to the full subgraph $\gS$. However, $\gG$ is generally too big to fit in GPU memory, 
so we sample the $k$-hop neighbors $\gN^k(\sV^{\obs})$ of the observed nodes.
There are two kinds of nodes in $k$-hop neighbors: $\sV^{\sub_k}$, nodes that are actually included in the full subgraph, and $\sV^{\glob_k}$, nodes that are not. Formally, 
\vspace{-0.1cm}
\begin{align*}
\sV^{\sub_k} = \gN^k(\sV^{\obs}) \cap \sV^{\sub},\ 
\sV^{\glob_k} = \gN^k(\sV^{\obs}) \cap (\sV^{\glob} \setminus \sV^{\sub}).
\end{align*}
Using the GD estimator, $k$-hop PSI maximizes MI between representations of $\sV^{\sub}$ and $\vs^{\obs}$ by using nodes in $\sV^{\sub_k} (\subset \sV^{\sub})$ as positive samples and nodes in $\sV^{\glob_k} (\subset \sV^{\glob})$ as negative samples.

The score $\gD (\vh_{v}, \vs^{\obs})$ in the GD estimator (Eq.~\ref{eq:infomax_loss}) can be interpreted as the probability that the node belongs to the subgraph. 
Using scores $\vd^{\ \khop} = \gD ( \mH^{\khop}, \vs^{\obs} )$ where $\mH^{\khop}$ are representations of $\gN^k(\sV^{\obs})$, we create $\vs^{\ \khop}$ for the final prediction,
\vspace{-0.37cm}
\begin{align}
\scale[0.94]{
\vs^{\ \khop} = \softmax (\vd^{\ \khop}_{[\idx]}) \cdot \MLP( \mH^{\khop}_{[\idx]} ),\ \idx = \topk ( \vd^{\ \khop} ),
}
\end{align}
a weighted average of $k$-hop neighbors by top-k values of $\vd^{\khop}$, where $\idx$ are top-k node indices. 
This $k$-hop PSI's pooling can be seen as a full subgraph reconstruction from the corrupted subgraph.

Although a summary from $k$-hop PSI is close to the full subgraph, its objective only relies on $k$-hop neighborhoods of the partial subgraph. For positive samples $\sV^{\sub_k}$, we cannot use the entire set of nodes $\sV^{\sub}$ in the full subgraph if $\sV^{\sub_k} \neq \sV^{\sub}$. In addition, the similar set of negative samples within $k$-hop will be drawn for each training iteration. To overcome this limitation, we propose two-stage models that link $k$-hop PSI with other PSI models. First, $k$-hop PSI produces the InfoMax loss $\gL^{\khop}$ and the summary $\vs^{\khop}$. We adopt the second PSI model, which uses this summary as input to distinguish positive and negative samples. Then, we compute the second InfoMax loss $\gL^{\textrm{2nd}}$. Finally, we jointly minimize all losses including $\gL^{\textrm{graph}}$, that is, $\gL^{\textrm{graph}} + \lambda^{\khop} \gL^{\khop} + \lambda^{\textrm{2nd}} \gL^{\textrm{2nd}}$.

\vspace{-0.21cm}\section{Experiments}\label{sec:experiments}
\renewcommand{\arraystretch}{0.9}

\begin{table}[]
\centering
\caption{Statistics of real-world datasets.}
\label{tab:dataset_statistics}
\vspace{-0.42cm}
{\small
\begin{tabular}{llll}
\hline
                            & FNTN            & EM-User             & HPO-Metab           \\ \hline
\# Global nodes             & 362,232    & 57,333   & 14,587               \\
\# Global edges             & 22,918,295 & 4,573,417 & 3,238,174             \\
Density of $\gG$ & 0.0002              & 0.0028              & 0.0304              \\ \hline
\# Subgraphs                & 1107                & 319                 & 2397                \\
\# Nodes per subgraph       & 408.6 $\pm$ 386.7 & 155.4 $\pm$ 100.4 & 14.4 $\pm$ 6.2    \\
\# Edges per subgraph       & 412.9 $\pm$ 391.3 & 534.9 $\pm$ 645.3 & 181.3 $\pm$ 181.8 \\
Density of $\gS$        & 0.004 $\pm$ 0.003 & 0.016 $\pm$ 0.005 & 0.758 $\pm$ 0.149 \\ \hline
\# Classes                  & 4                   & 2                   & 6                   \\ \hline
\end{tabular}
}
\vspace{-0.4cm}
\end{table}

\vspace{-0.1cm}
\paragraph{Datasets}\label{sec:datasets}

We experiment with three real-world datasets. 
\FNTNb (Fake News in Twitter Network; ordered) \citep{liu2015real, ma2016detecting, ma2017detect, ma2018rumor, kim2019homogeneity} is a Twitter network ($\gG$) with news propagation trees ($\sS$), contents ($\vg$), and genuineness ($y$). The fake news early detection task is classifying the genuineness of news by initial nodes. \EMUserb (Users in EndoMondo; unordered) \citep{ni2019endomondo, alsentzer2020subgraph} is a fitness network of workouts ($\gG$), users as subgraphs ($\sS$), and their gender ($y$), where the task is to profile a user's gender with only a few logs. The global graph $\gG$ of \HPOMetabb (Metabolic disease in Human Phenotype Ontology; unordered) \citep{hartley2020new, kohler2019expansion, mordaunt2020metabolomics, alsentzer2020subgraph} is a knowledge graph of symptoms. Each subgraph $\gS$ is a collection of symptoms associated with a metabolic disease, and the label ($y$) is the disease type. The task is to classify the disease type, assuming only some of the symptoms are observed. Detailed statistics are reported in Table~\ref{tab:dataset_statistics}. We randomly split the train/val/test set of \FNTNb with a ratio of 70/15/15 and use public splits~\citep{alsentzer2020subgraph} for \EMUserb (70/15/15) and \HPOMetabb (80/10/10).

\vspace{-0.21cm}
\paragraph{Training and evaluation settings}\label{sec:experimental_set_up}

For both training and evaluation, we set the number of observed nodes $|\sV^{\obs}|$ to 4 for \HPOMetab, (the average number of nodes $<$ 16), and 8 for \FNTNb and \EMUser, (the average size of subgraphs $\gg$ 16). Further, to see the performance change with $|\sV^{\obs}|$, we conduct experiments where $|\sV^{\obs}|$ is 8, 16, 32, and 64 for \FNTNb and \EMUser. We also experiment with an oracle setting where all subgraphs are fully observed.

\vspace{-0.21cm}
\paragraph{Model and training details}

For the encoder $\gE$, we use the two-layer GraphSAGE~\citep{hamilton2017inductive} with skip connections~\citep{he2016deep}. As an input of $\gE$, node features $\mX^{\glob} \in \sR^{|\sV^{\glob}| \times F^{\textrm{in}}}$ are trainable parameters with $F^{\textrm{in}}$ of 32 (\FNTN) and 64 (others). For \HPOMetabb and \EMUser, we use pre-trained embeddings from \citet{alsentzer2020subgraph}. 
We use readout $\gR$ of mean pooling after a two-layer MLP for all models except for $k$-hop PSI in the two-stage model, where we use the soft-attention pooling~\citep{li2015gated} after a two-layer Transformer~\citep{vaswani2017attention}.
We add the positional encoding before Transformer for ordered \FNTN. For the discriminator $\gD$, we use a bilinear scoring~\citep{velickovic2018deep, oord2018representation} for the GD estimator, and cosine similarity with a temperature~\citep{you2020graph} for the InfoNCE estimator. For the prediction function $\gF$, we use a single-layer neural network. If there is a subgraph-level feature $\vg \in \sR^{F'}$, we first transform it to the vector of the same length as $\vs$, concatenate it with $\vs$, and feed them to the prediction layer. All models use $F=64$ features, the ReLU activation, dropout of 0.2~\citep{srivastava2014dropout}, and the Adam optimizer~\citep{kingma2014adam} with a learning rate of $10^{-3}$. We sample nodes in a one-hop neighborhood in the $k$-hop PSI (i.e., $k=1$). They are implemented with PyTorch ecosystems~\citep{paszke2019pytorch, Fey2019Fast, Zhu2021tu, Falcon_PyTorch_Lightning_2019}.


\vspace{-0.21cm}
\paragraph{Two-stage models} 

We use $k$-hop PSI + PS-DGI and + PS-InfoGraph only for two-stage models since using PS-MVGRL or PS-GraphCL as a second model is practically difficult. For MVGRL, using non-shared encoders and PPR augmentation requires significant memory and computations, and for GraphCL, a large batch size is needed.

\vspace{-0.21cm}
\paragraph{Baselines}

All baselines share the following encoder-readout architecture: $\mH = \gE^B (\mX[\sV^{\obs}], \sA^{\obs}),\ 
\vs^{B} = \gR^B (\mH),\ 
\vy = \gF(\vs^{B}, [\vg]),$ where $\gE^B$ are two-layer MLP, GCN~\citep{kipf2017semi}, GraphSAGE~\citep{hamilton2017inductive}, GAT~\citep{velickovic2018graph}, and SubGNN~\citep{alsentzer2020subgraph} with skip connections. We set $\gR^B$ the two-layer MLP after mean pooling for SubGNN and mean pooling after two-layer MLP for others. We report the performance of SubGNN on \EMUserb and \HPOMetabb only since SubGNN requires large memory of $O(|\sV^{\glob}|^2)$ (> 1TB for \FNTN) in computing shortest paths.

\vspace{-0.21cm}
\section{Results and Discussions}\label{sec:results}
\renewcommand{\arraystretch}{0.95}

\begin{table}[]
\centering
\caption{
\fontsize{8.87}{11}\selectfont 
Mean and standard deviation of accuracy of five runs. The first column is the number of observed nodes $ | \sV^{\obs} | $ (\S\ref{sec:experimental_set_up}), and the setting of each dataset is indicated with $\dag, \ddag$. \colorbox{weakblue}{The PSI model} that outperforms \colorbox{weakred}{the best baseline} (except for \colorbox{weakgray}{the oracle}) is indicated by color, and statistical significance by unpaired \ttest\ by asterisks (\colorbox{midblue}{**$p < .001$, *$p < .1$}).
}
\vspace{-0.375cm}
\label{tab:main_result}
{\small
\begin{tabular}{lllll}
\hline
$ | \sV^{\obs} | $    & Model        & FNTN$^{\ddag}$             & EM-User$^{\ddag}$ & HPO-Metab$^{\dag}$   \\ \hline
100\%                 & GraphSAGE    & \cellwg$86.3_{\pm 0.7}$    &\cellwg$82.1_{\pm 1.2}$ &\cellwg$47.7_{\pm 3.3}$ \\ \hline
                      & MLP          & $82.5_{\pm 2.6}$           & $71.9_{\pm 4.6}$       & $43.5_{\pm 4.4}$       \\
                      & GCN          & $84.6_{\pm 2.0}$           &\cellwr$72.8_{\pm 3.8}$ & $42.9_{\pm 1.8}$       \\
                      & GraphSAGE    & $84.9_{\pm 1.3}$           & $68.1_{\pm 2.6}$       &\cellwr$44.1_{\pm 1.3}$ \\
                      & GAT          &\cellwr$85.1_{\pm 0.8}$     & $71.5_{\pm 5.7}$       & $43.1_{\pm 2.3}$       \\
                      & SubGNN       & N/A                        & $61.3_{\pm 5.4}$       & $37.1_{\pm 1.5}$       \\ \cline{2-5} 
                      & PS-DGI       &\cellmb$87.5_{\pm 1.2}^{*}$ & $72.3_{\pm 6.2}$       & $44.0_{\pm 1.8}$       \\
$8^{\ddag}, 4^{\dag}$ & PS-InfoGraph &\cellmb$87.3_{\pm 0.0}^{**}$&\cellwb$75.7_{\pm 3.9}$ &\cellmb$47.1_{\pm 2.1}^{*}$   \\
                      & PS-MVGRL     &\cellmb$88.6_{\pm 0.9}^{**}$& OOM                    &\cellwb$45.4_{\pm 2.4}$       \\
                      & PS-GraphCL   &\cellmb$88.1_{\pm 1.3}^{*}$ &\cellwb$75.3_{\pm 2.4}$ &\cellmb$47.2_{\pm 3.5}^{*}$   \\
                      & $k$-hop PSI  &\cellmb$87.8_{\pm 1.2}^{*}$ &\cellwb$75.3_{\pm 2.4}$ & $42.4_{\pm 2.6}$      \\ \cline{2-5} 
                      & \begin{tabular}[c]{@{}l@{}}$k$-hop PSI\\[-0.034cm] \ + PS-DGI\end{tabular}  &\cellmb$88.0_{\pm 0.7}^{**}$ &\cellwb$75.7_{\pm 4.4}$     &  $43.6_{\pm 1.0}$  \\
                      & \begin{tabular}[c]{@{}l@{}}$k$-hop PSI\\[-0.034cm] \ + PS-InfoGraph\end{tabular} &\cellmb$89.6_{\pm 1.8}^{**}$ &\cellmb$77.0_{\pm 5.2}^{*}$ &\cellwb$44.6_{\pm 1.6}$ \\ \hline
\end{tabular}
}
\vspace{-0.5cm}
\end{table}


\vspace{-0.1cm}
\paragraph{Performance by models and datasets}

Table~\ref{tab:main_result} summarizes the mean accuracies over five runs of various models. We confirm that PSI models outperform all comparison models for all three datasets except for PS-DGI and $k$-hop PSI. PS-DGI performs worse than the best baseline in \EMUserb and \HPOMetab. Among PSI models, PS-InfoGraph and PS-GraphCL consistently outperform baselines across datasets. PS-MVGRL shows the best performance among single PSI models on \FNTN, but it does not fit in single GPU (VRAM of $11G$) on \EMUser. However, the performance differences among the PSI models except for $k$-hop PSI are not statistically significant ($\textrm{\pvalue} > .1$ for all datasets in one-way ANOVA).

In \FNTNb and \EMUser, $k$-hop PSI is on par with other PSI models, but in \HPOMetab, $k$-hop PSI significantly underperforms. This behavior is caused by differences in the density of the global graph. As in Table~\ref{tab:dataset_statistics}, \HPOMetabb has a higher density ($0.03$) than \FNTNb ($2 \times 10^{-4}$) and \EMUserb ($2.8 \times 10^{-3}$). When $k$-hop subgraph sampling is used, more neighbor nodes are included for denser graphs. Since most of the sampled neighbors are not in the subgraph, it is difficult to distinguish which of many neighbors belong to the subgraph by the discriminator $\gD$ in $k$-hop PSI. It degrades the performance of discriminator $\gD$ and the classification performance eventually.

Two-stage model's performance is mainly driven by the $k$-hop PSI's performance. In \HPOMetabb where $k$-hop PSI does not perform well, the performance is lower than single PSIs; otherwise, it outperforms single models. The noise from high density is still relevant in two-stage models. In all datasets, a two-stage model results in better performance than a single $k$-hop PSI, and the combination with PS-InfoGraph performed better than with PS-DGI.

Finally, we discuss the results of baselines. First, SubGNN underperforms simple models. SubGNN uses message-passing between subgraphs, and thus partial observation degrades its performance. Second, there is no significance difference in MLP, GCN, GraphSAGE, and GAT ($\textrm{\pvalue} > .1$ in one-way ANOVA).

\vspace{-0.21cm}
\paragraph{Performance by the number of observed nodes}

\begin{figure}
\centering
  \begin{subfigure}[t]{0.207\textwidth}
    \centering
    \includegraphics[height=0.12\textheight]{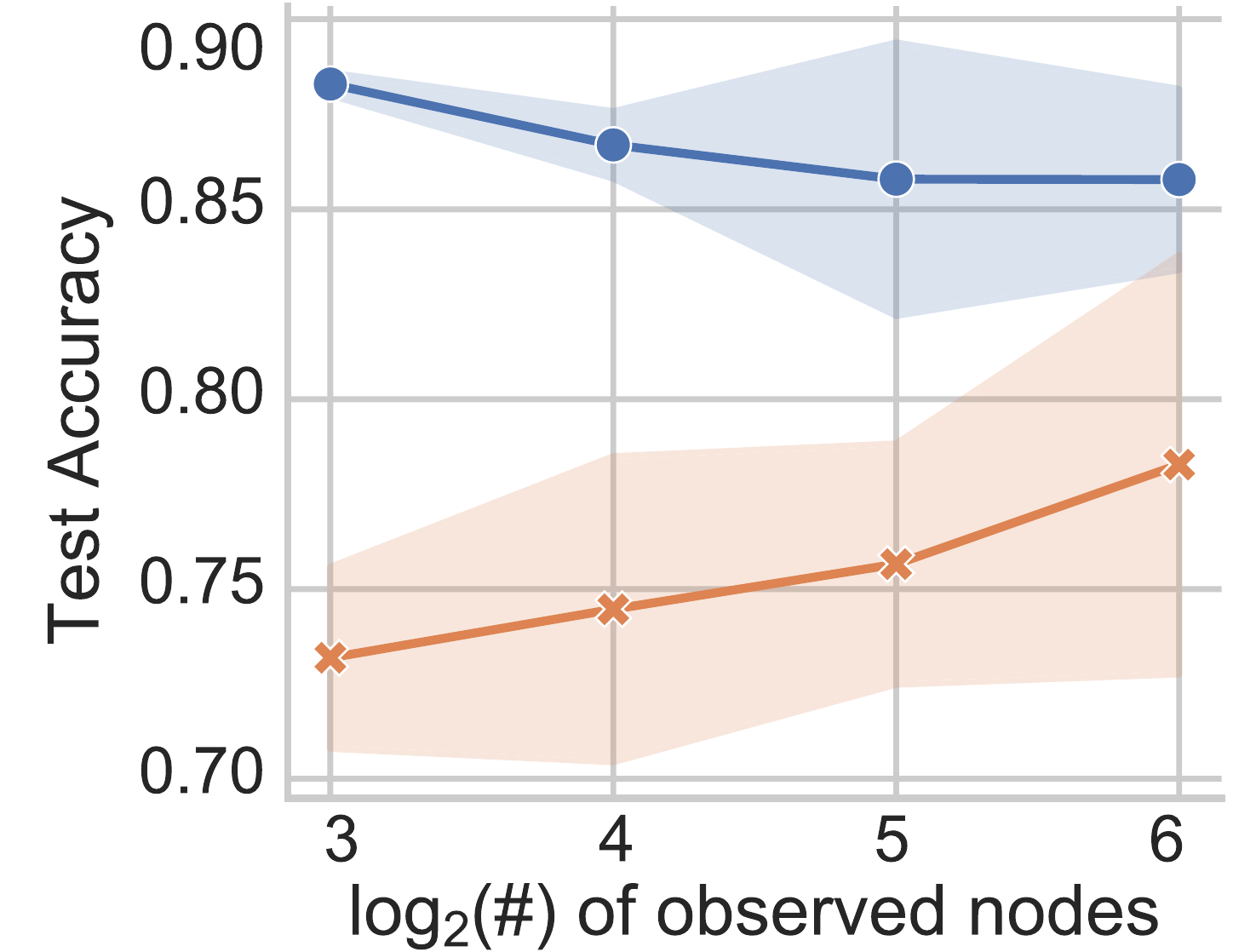}
    \vspace{-0.19cm}
    \caption{At training and test stages.}
    \label{fig:num_obs_xx}
  \end{subfigure}
  \begin{subfigure}[t]{0.202\textwidth}
    \centering
    \includegraphics[height=0.12\textheight]{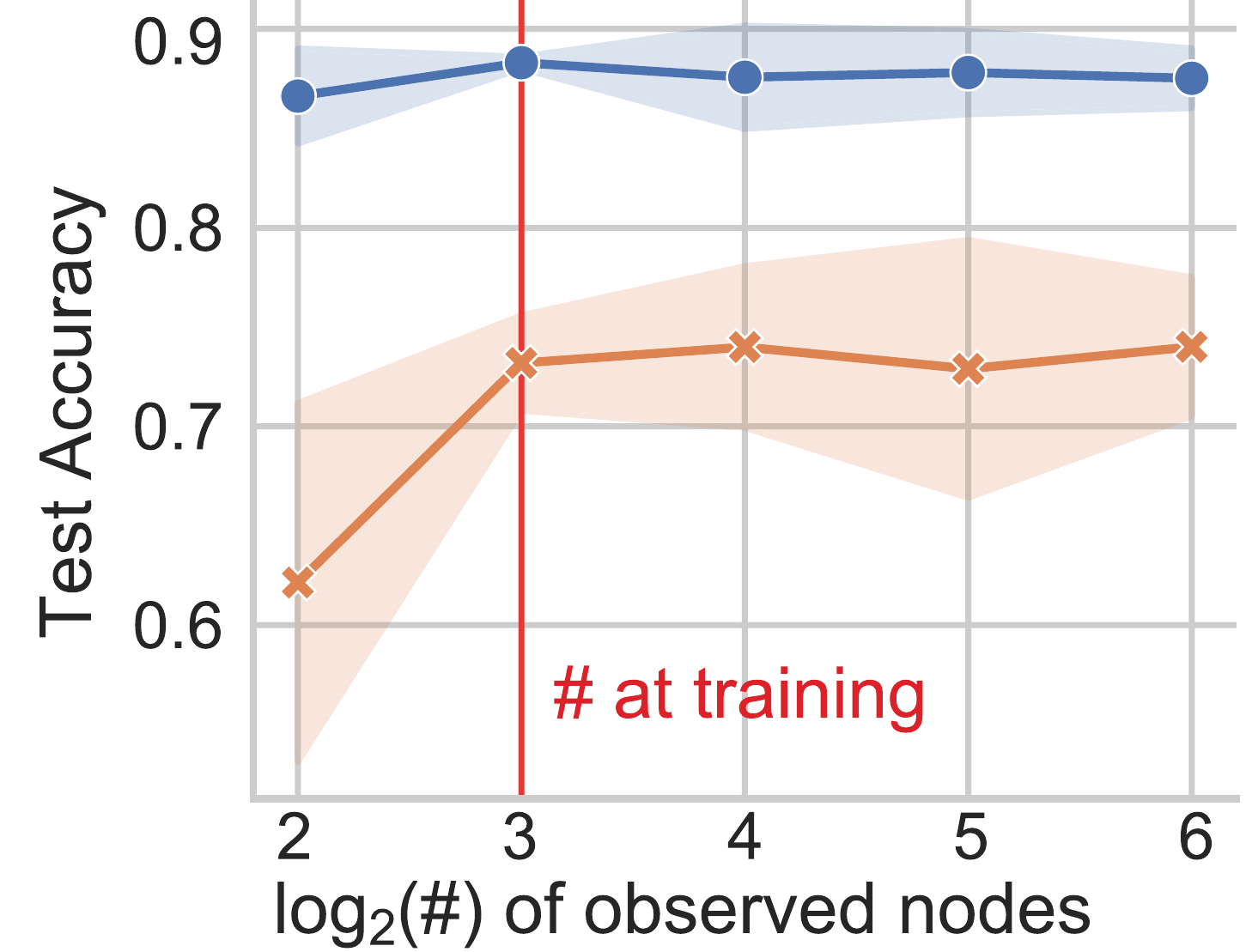}
    \vspace{-0.19cm}
    \caption{At the test stage.}
    \label{fig:num_obs_eval}
  \end{subfigure}
  \begin{subfigure}[t]{0.05\textwidth}
    \centering
    \includegraphics[height=0.11\textheight]{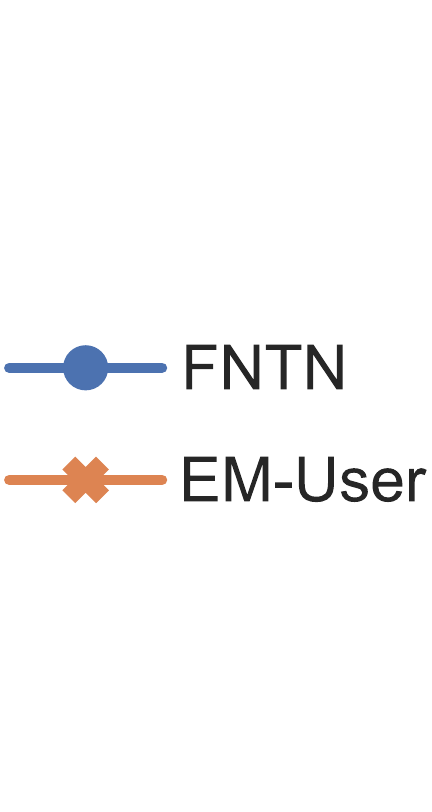}
  \end{subfigure}
  \vspace{-0.39cm}
  \caption{Performance of the $k$-hop PSI + PS-InfoGraph by the number of observed nodes at specific stages.}
  \label{fig:num_obs_all}
  \vspace{-0.5cm}
\end{figure}

In Figure~\ref{fig:num_obs_xx}, we show the mean accuracy of $k$-hop PSI + PS-InfoGraph (5 runs) by the number of observed nodes in training and test. We exclude \HPOMetabb with an average number of nodes fewer than 64. Intuitively, more observations should result in better prediction, and the performance on \EMUserb is consistent with that intuition. However, for \FNTN, the opposite is true because initial nodes are relatively important for the propagation-based fake news detection~\citep{bian2020rumor}. Note that adding observed nodes is equivalent to adding $\khop$ neighbors to be discriminated by $\gD$ in $k$-hop PSI. That is, the impact of performance degradation from neighborhood noise is more significant than information gain from additional nodes in \FNTN.

\vspace{-0.21cm}
\paragraph{Generalization across sizes of \textit{test} subgraphs}

Figure~\ref{fig:num_obs_eval} shows how $k$-hop PSI + PS-InfoGraph generalizes across sizes of test subgraphs (mean performance over 5 runs). We set the number of \textit{test} observed nodes from 4 to 64 and fix the number of \textit{training} observed nodes to 8. Our model generalizes on test samples with more observed nodes ($>8$) than training, but the variance of performances increases. In contrast, there is a lack of generalizability for test samples with fewer observed nodes than in the training stage. In particular, some trials do not converge on \EMUser.

\begin{figure}
\centering
  \begin{subfigure}[t]{0.195\textwidth}
    \centering
    \includegraphics[height=0.124\textheight]{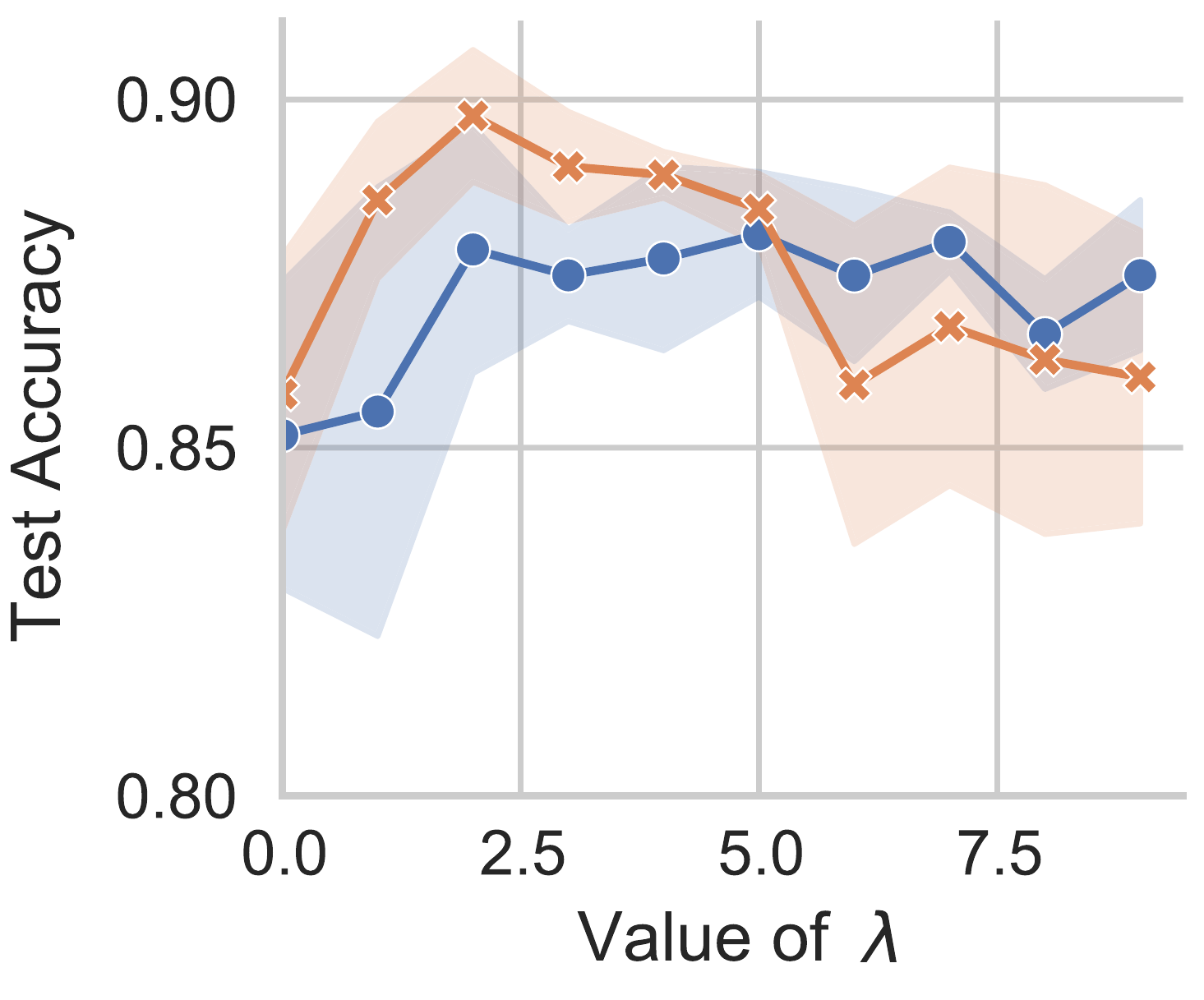}
    \vspace{-0.19cm}
    \caption{FNTN}
    \label{fig:sensitivity-fntn}
  \end{subfigure}
  \begin{subfigure}[t]{0.195\textwidth}
    \centering
    \includegraphics[height=0.124\textheight]{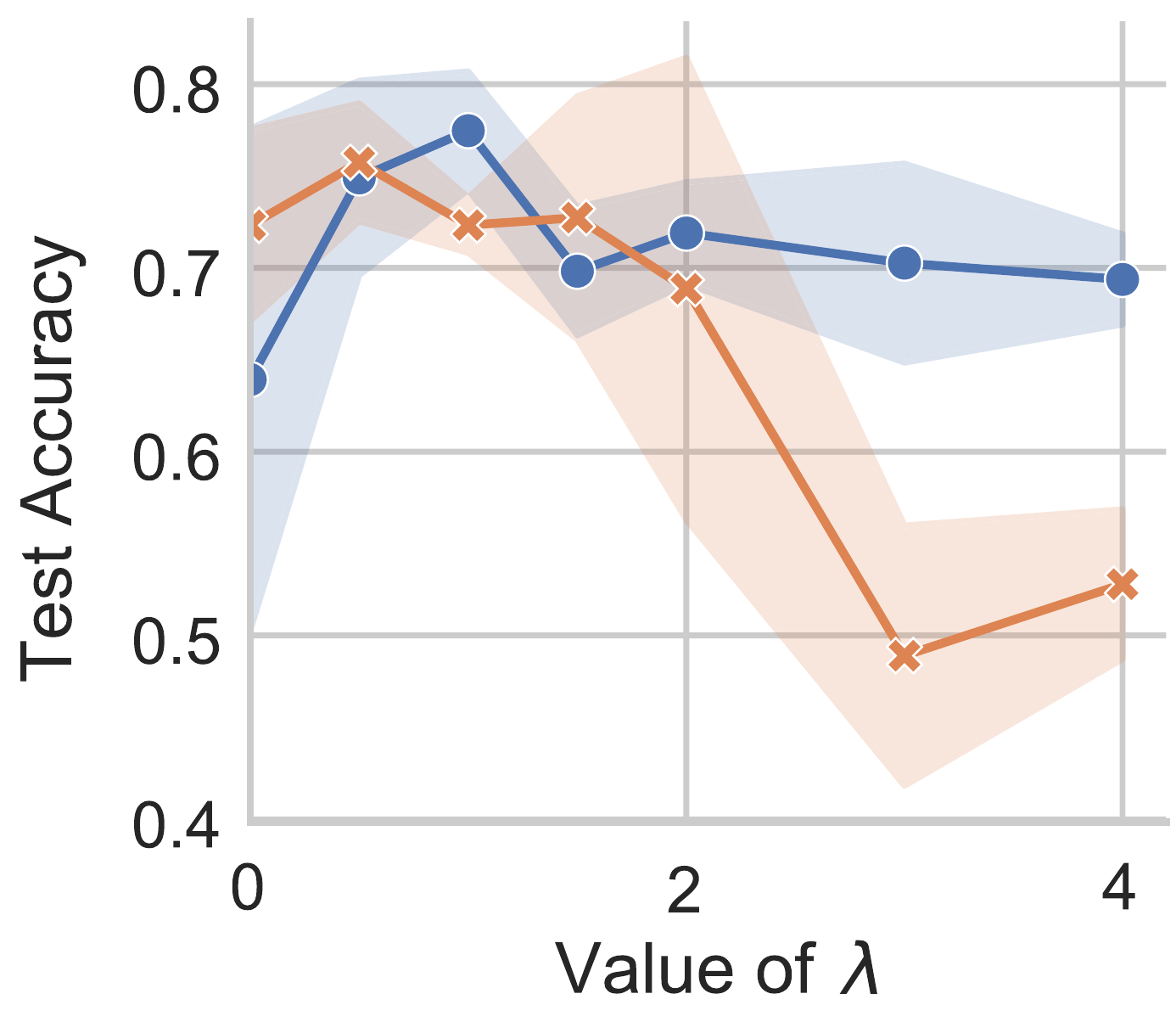}
    \vspace{-0.19cm}
    \caption{EM-User}
    \label{fig:sensitivity-emuser}
  \end{subfigure}
  \begin{subfigure}[t]{0.05\textwidth}
    \centering
    \includegraphics[height=0.11\textheight]{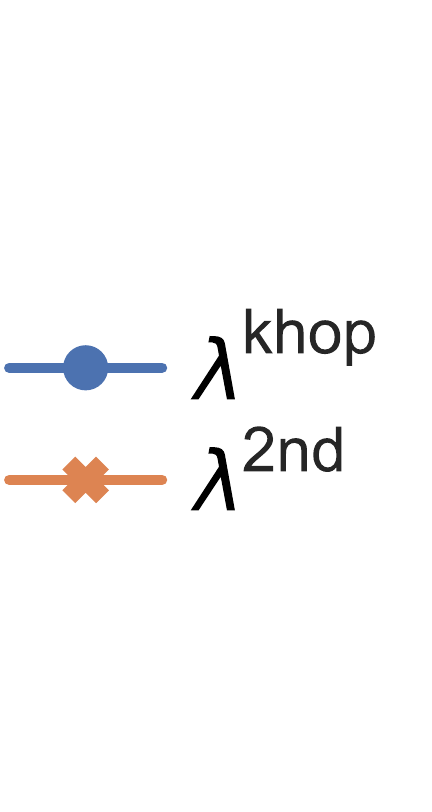}
  \end{subfigure}
  \vspace{-0.39cm}
  \caption{Mean and standard deviation of test accuracy (5 runs) on \FNTNb and \EMUserb against $\lambdakhop$ and $\lambda^{\textrm{2nd}}$.}
  \label{fig:sensitivity}
  \vspace{-0.52cm}
\end{figure}

\vspace{-0.21cm}
\paragraph{Sensitivity to $\lambda$}

In Figure~\ref{fig:sensitivity}, we plot the test accuracy on \FNTNb and \EMUserb against $\lambdakhop$ (with $k$-hop PSI) and $\lambda^{\textrm{2nd}}$ (with $k$-hop PSI + PS-InfoGraph). The sensitivity to $\lambda^{\textrm{2nd}}$ is higher than $\lambdakhop$ in both datasets. Notably, in the region where $\lambdakhop$ is larger than the optimum, the performance fluctuates slightly less. A large $\lambdakhop$ makes the discriminator $\gD$ of $k$-hop PSI overestimate the probability of belonging to the subgraph. Since we only use a fixed ratio in the $\topk$ pooling, those classified as nodes belonging to the subgraph more than this ratio are not involved in subgraph representations, thus no significant change where $\lambdakhop$ is large.



\vspace{-0.19cm}
\section{Conclusion}\label{sec:conclusion}
\vspace{-0.08cm}
We explored the `partial subgraph learning task' where only a part of the subgraph is observed. This is a more realistic and challenging scenario of subgraph representation learning. We also proposed a novel framework, Partial Subgraph Infomax (PSI), which maximizes the mutual information between the partial subgraph's summary and representations of substructures like nodes or full subgraphs. Using training and evaluation protocols designed to simulate real-world use cases, PSI models outperform baselines in three datasets. One limitation is that $k$-hop PSI uses a naive $\khop$ sampling to select neighbors to be included in the subgraph, which is a major cause of performance degradation in dense graphs. We leave how to effectively and efficiently choose nodes as future work.


\begin{acks}
This research was supported by the ERC Program through the NRF funded by the Korean Government MSIT (NRF-2018R1A5A1059921).
\end{acks}

\clearpage

\bibliographystyle{ACM-Reference-Format}
\balance
\bibliography{sample-base}

\def\notappendix{0} 

\if\notappendix1
\else
\clearpage
\appendix
\appendix

The Appendix is not peer-reviewed and not a part of the proceedings of the 31st ACM International Conference on Information and Knowledge Management.

\section{Related Work}
Our study tackles a generalized subgraph representation learning with contrastive learning by mutual information maximization. This section introduces these two research areas.

\vspace{-0.16cm}
\subsection{Subgraphs in graph representation learning}


There have been several approaches to use the information in subgraphs to improve representation learning of graph-structured data. They employ subgraphs to build more expressive models for node and graph representations~\citep{niepert2016learning, bouritsas2020improving, yu2021graph}, improve the scalability of graph neural network (GNN) training~\citep{hamilton2017inductive, chiang2019cluster, Zeng2020GraphSAINT:}, augment data for graphs~\citep{qiu2020gcc, you2020graph}, and explain prediction results of GNNs~\citep{ying2019gnnexplainer, luo2020parameterized}.
Another common approach is to learn (or meta-learn) nodes or edges of interest by fetching a local (or enclosing) subgraph around them~\citep{bordes2014question, zhang2018link, teru2020inductive, huang2020graph}.

While these methods target node- or graph-level tasks, a few studies focus on the subgraph-level task. \citet{meng2018subgraph} classifies the subgraph evolution pattern for subgraphs induced by three or four nodes as inputs, but it does not learn the representation of subgraphs. Subgraph Neural Network (SubGNN)~\citep{alsentzer2020subgraph} is designed for subgraph-level classification with subgraph representation learning using their internal/external topology, positions, and connectivity. SubGNN assumes that all subgraphs are fully observed, whereas our model does not make this assumption.

\vspace{-0.16cm}
\subsection{Contrastive learning by mutual information maximization}

Contrastive learning is a widely-used method for self- and un-supervised learning~\citep{liu2020self, le2020contrastive}. This has been applied in various types of data such as language, nodes, and images. Within contrastive learning, InfoMax methods~\citep{hjelm2018learning} have been proposed recently, leveraging the known structure of data while maximizing mutual information  (MI) of input and encoded output. Specifically, they maximize MI between pairs of local (e.g., patches) and global (e.g., images) based on neural MI estimators~\citep{belghazi2018mutual, nowozin2016fgan, oord2018representation}. 

For graphs, various inherent substructures can be used in the design of contrastive learning. For example, node representations can be obtained by maximizing MI between node-graph pairs~\citep{velickovic2018deep, park2020unsupervised, hassani2020contrastive, wang2020exploiting, jing2021hdmi}, node-subgraph pairs~\citep{peng2020graph, jiao2020sub, li2020graph}, edge-edge pairs~\citep{peng2020graph}, subgraph-graph pairs~\citep{cao2021bipartite}. Likewise, graph representations can be obtained by maximizing MI between node-graph pairs~\citep{Sun2020InfoGraph:, hassani2020contrastive}, node-subgraph pairs~\citep{li2020graph}, subgraph-graph pairs~\citep{sun2021sugar}, and graph-graph pairs~\citep{you2020graph}. Our model families learn representations of partial subgraphs by maximizing MI with other substructures like nodes or full subgraphs. To the best of our knowledge, there is no InfoMax method designed to learn subgraph representation itself, regardless of the conditions of incomplete observations.

\section{Dataset Pre-processing Details}\label{appendix:datasets}

For \FNTN, a follower network was crawled through the Twitter API between October and November 2018 for users in the propagation trees (including leaf users)~\citep{liu2015real, ma2016detecting, ma2017detect, ma2018rumor, kim2019homogeneity}. For deactivated accounts, we reflect the following information that can be obtained from the tree. We collect and distribute these data under Twitter's policies and agreements\footnote{\url{https://developer.twitter.com/en/developer-terms}}.

Datasets in this paper are pre-processed to remove any personally identifiable information of users in real-world services (Twitter for \FNTNb and Endomondo for \EMUser). Users are fully anonymized and treated as consecutive integers. In addition, we take TF-IDF vectors of 2000 words for news content without stop-words. The fake news texts, which can be offensive, cannot be restored.

For all datasets, single node graphs are excluded (five subgraphs for \EMUserb and three subgraphs for \HPOMetab). The rests are the same as the original papers (See \citet{kim2019homogeneity} for \FNTN, and \citet{alsentzer2020subgraph} for \EMUserb and \HPOMetab).

The raw datasets of \HPOMetabb and \EMUserb can be downloaded from SubGNN's GitHub repository\footnote{\url{https://github.com/mims-harvard/SubGNN}}. The codes using Twitter API to construct \FNTNb are public in the GitHub Repository\footnote{\url{https://github.com/dongkwan-kim/Fake-News-Twitter-Network}}.

\section{Details of the Partial Subgraph InfoMax Framework}

In addition to the description in the Table~\ref{tab:psi_models}, we give more details of the PSI framework and models in it. We generalize existing InfoMax models for learning partial subgraph representations. We consider models that can maximize MI between local and global structures in the graph, including DGI~\citep{velickovic2018deep}, InfoGraph~\citep{Sun2020InfoGraph:}, MVGRL~\citep{hassani2020contrastive}, and GraphCL~\citep{you2020graph}. Since they are initially designed for the node or graph-level prediction, we incorporate them into the PSI framework by revising the core steps of the encoder-readout pipeline. Specifically, we view each partial subgraph as an independent graph and generate its summary vector as described in steps 1 -- 3 of Algorithm~\ref{alg:psi}. We maximize the MI between the partial subgraph summary and other substructures in full subgraphs (step 4). We follow the original models for the rest of the architecture.

\begin{figure*}[t]
  \centering
  \includegraphics[width=0.99\textwidth]{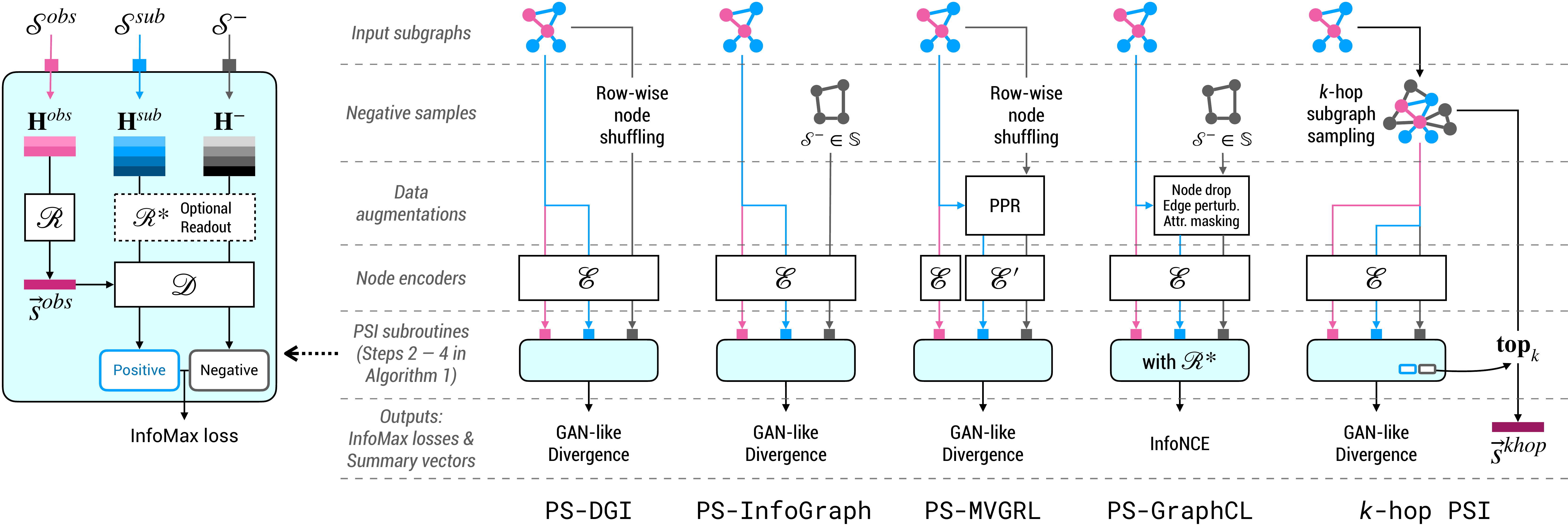}
  \vspace{-0.25cm}
  \caption{Models in the PSI framework: PS-DGI, PS-InfoGraph, PS-MVGRL, PS-GraphCL, and $k$-hop PSI.}
  \vspace{-0.2cm}
  \label{fig:model}
\end{figure*}


\cmt{

\begin{figure*}[t]
  \centering
  \begin{subfigure}[b]{0.792\textwidth}
      \centering
      \includegraphics[width=\textwidth]{figures/PSI_model_1.pdf}
      \caption{Models in the PSI framework with different InfoMax losses, negative sampling, and augmentations: PS-DGI, PS-InfoGraph, PS-MVGRL, PS-GraphCL, and $k$-hop PSI. The PSI block is a subroutine that discriminates (by $\gD$) the partially observed subgraph summary $\vs^{\obs}$ (by $\gR$) from positive ($\gS^{\sub}$) and negative ($\gS^{-}$) substructures. For PS-GraphCL, these substructures are full subgraphs (summarized by $\gR^{*}$); otherwise, they are nodes.}
      \label{fig:model_individual}
  \end{subfigure}
     \hfill
  \begin{subfigure}[b]{0.187\textwidth}
    \centering
    \includegraphics[width=\textwidth]{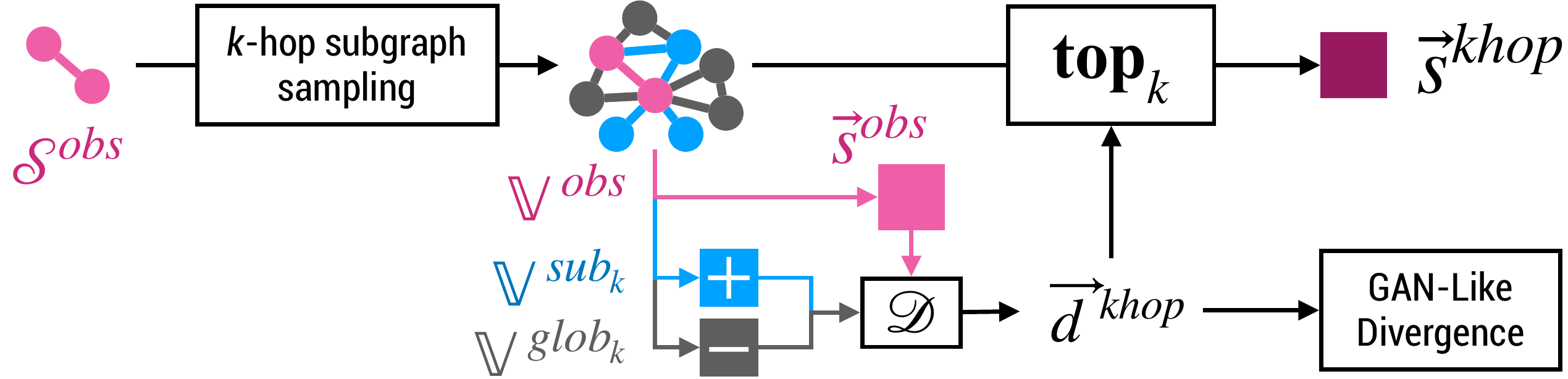}
    \caption{Two-stage PSI models. The reconstructed summary $\vs^{\khop}$ from $k$-hop PSI distinguishes positive and negative substructures. Then, two InfoMax losses are jointly optimized.}
    \label{fig:model_two_stage}
  \end{subfigure}
  \vspace{-0.25cm}
  \caption{Visual summaries of the PSI framework: individual (\ref{fig:model_individual}) and two-stage (\ref{fig:model_two_stage}) models.}
  \label{fig:model}
  \vspace{-0.21cm}
\end{figure*}

}

\begin{algorithm}
\caption{Partial Subgraph InfoMax framework}\label{alg:psi}
\KwData{The global graph $\gG = (\sV^{\glob}, \sA^{\glob}), \mX^{\glob}$ and a partially observed subgraph $\gS^{\obs}$ in $\sS^{\obs}$: $\gS^{\obs} = (\sV^{\obs}, \sA^{\obs}), \vg, \mX^{\obs} = \mX^{\glob}[\sV^{\obs}]$}
\KwResult{A logit vector of $\gS^{\obs}$: $\vy \in \sR^{C}$.}

1. Encode nodes in the partial subgraph: $\mH^{\obs} = \gE ( \mX^{\obs}, \sA^{\obs} )$.

2. Create the partial subgraph summary: $\vs^{\obs} = \gR (\mH^{\obs})$.

3. Compute a logit vector: $\vy = \gF (\vs^{\obs}, [\vg])$. We also use a subgraph-level feature $\vg$ in the final prediction if exists.

4. In the training stage, optimize losses to update parameters:

\If{training}
{
    a. Draw negative subgraph(s): $\gS^{-} = (\sV^{-}, \sA^{-}) \in \sS$.
    
    b. Augment full and negative subgraphs if necessary:
    
    \If{graph augmentations are used}{
        $\gS^{\sub}, \gS^{-} = \textrm{Aug}(\gS^{\sub}), \textrm{Aug}(\gS^{-})$.
    }
    
    c. Encode nodes in full and negative subgraphs, then summarize them if necessary:
    $\mH^{\sub} = \gE ( \mX[\sV^{\sub}], \sA^{\sub} \ \textor\ \sA^{\glob}[\sV^{\sub}] ),$
    $\mH^{-} = \gE ( \mX[\sV^{-}], \sA^{-} \ \textor\ \sA^{\glob}[\sV^{-}] ),$
    
    \ \ $\vs^{\sub},\ \vs^{\ -} = \gR (\mH^{\sub}),\ \gR (\mH^{-})$.
    
    d. Compute $\gL^{\textrm{GD}\ \textor\ \textrm{InfoNCE}}$ with the partial subgraph summary ($\vs^{\obs}$), positive ($\mH^{\sub}$ or $\vs^{\sub}$) and negative ($\mH^{-}$ or $\vs^{\ -}$) samples (Equations~\ref{eq:infomax_loss} or \ref{eq:infonce_loss}).
    
    e. Compute the cross-entropy loss $\gL^{\textrm{graph}}$ on the logit $\vy$ and label $y$, and minimize $\gL^{\textrm{graph}} + \lambda \gL^{\textrm{GD}\ \textor\ \textrm{InfoNCE}}$.
}
\end{algorithm}

A visual summary of PSI models (PS-DGI, PS-InfoGraph, PS-MVGRL, PS-GraphCL, and $k$-hop PSI) is illustrated in Figure~\ref{fig:model}. We focus on (1) what substructure pairs are used to maximize MI, (2) how negative samples are drawn from $\Tilde{P}$, and (3) what graph augmentation methods are used.

\section{Discussion on negative sampling in $k$-hop PSI}

Using the GAN-like divergence estimator (Equation~\ref{eq:infomax_loss}), $k$-hop PSI maximizes the MI between representations of $\sV^{\sub}$ and $\vs^{\obs}$ by using nodes in $\sV^{\obs} \cup \sV^{\sub_k}$ as positive samples and nodes in $\sV^{\glob_k}$ as negative samples, that is,
\begin{align}
\scale[0.906]{
\begin{aligned}
\gL^{\khop} = \textstyle
\frac{1}{|\sV^{\obs} \cup \sV^{\sub_k} \cup \sV^{\glob_k}|} \Big[
& \textstyle \sum_{v \in \sV^{\obs} \cup \sV^{\sub_k}}
    \log \sigma \left(
        \gD ( \vh_{v}, \vs^{\obs} )
    \right) + \\
& \textstyle \sum_{\Tilde{v} \in \sV^{\glob_k}}
    \log \left(
        1 - \sigma \left(
            \gD ( \vh_{\Tilde{v}}, \vs^{\obs} )
        \right)
    \right)
\Big].
\end{aligned}
}
\label{eq:khop_psi_loss}
\end{align}

In $k$-hop PSI, the negative nodes $\sV^{\glob_k}$ are not sampled from the true marginal distribution. Intuitively, this $\sV^{\glob_k}$, a set of nodes closely linked to the subgraph within $k$-hop, can be considered hard negative samples conditioned on positive samples. This approach is known to learn a better representation in contrastive and metric learning~\citep{oh2016deep, oord2018representation, zhuang2019local}, but such non-i.i.d sampling may break the assumption on the MI bound~\citep{tschannen2020On}. In Proposition~\ref{prop:conditional-gan-mi}, similar to Conditional-NCE (CNCE)~\citep{wu2021conditional}, we prove that a specific choice of negative sample distribution forms the lower bound of the GAN-like divergence MI estimator.
\vspace{-0.13cm}
\propositioncondgan
\begin{proof}
It suffices to show that $\E_{y \sim p(y)} \left[ \log \left( 1 + e^ {f (x, y)} \right) \right] \leq
\E_{y \sim q(y \vert x)} \left[ \log \left( 1 + e^ {f (x, y)} \right) \right]$ for all $x$ to prove $\gI^{\textrm{CGD}} \leq \gI^{\textrm{GD}}$, since,
\begin{align}
&\E_{y \sim p(y)} \left[ \log \left( 1 + e^ {f (x, y)} \right) \right] \leq
\E_{y \sim q(y \vert x)} \left[ \log \left( 1 + e^ {f (x, y)} \right) \right] \label{eq:key_statement} \\
&\begin{aligned}
\Rightarrow\ &\E_{x \sim p(x), y \sim p(y)} \left[ \log \left( 1 + e^ {f (x, y)} \right) \right] \leq \\
&\E_{x \sim p(x), y \sim q(y \vert x)} \left[ \log \left( 1 + e^ {f (x, y)} \right) \right]
\end{aligned} \\
&\begin{aligned}
\Rightarrow\ & \E_{x \sim p(x), y \sim q(y \vert x)} \left[ \log \left( 1 - \sigma ( f (x, y)) \right) \right] \leq \\
&\E_{x \sim p(x), y \sim p(y)} \left[ \log \left( 1 - \sigma ( f (x, y)) \right) \right]
\end{aligned} \\
&\Rightarrow \gI^{\textrm{CGD}} \leq \gI^{\textrm{GD}}
\end{align}
We apply the similar technique in CNCE~\citep{wu2021conditional} to prove Equation~\ref{eq:key_statement}. Using Jensen's inequality to the right-hand side and the fact that $\E_p (e^{f(x, y)}) \leq e^{f(x, y_c)}$ for $y_c \in \sY_c$,
\begin{align}
\scale[0.875]{
\E_{p} \left[ \log \left( 1 + e^ {f (x, y)} \right) \right] \leq
\log \E_{p} \left[ \left( 1 + e^ {f (x, y)} \right) \right] \leq
\log \left( 1 + e^ {f (x, y_c)} \right).
}
\end{align}
If we take the expectation $\E_{y \sim q(y \vert x)}$ on both sides, we get Equation~\ref{eq:key_statement}.
\end{proof}
After applying $f$ to the training set, the CNCE uses a subset, the exponentiated similarity $e^{f(\cdot, \cdot)}$ of which is bigger than that of a certain percentile. Instead, we employ $k$-hop sampling, which uses hop distance as a proxy of embedding distance (or dissimilarity). This method assumes that the hop and embedding distances of nodes created by message-passing are highly correlated. It is more efficient than using the actual similarity since it does not evaluate $f$ for all instances.

\section{Model, training, and hyperparameter configurations}\label{appendix:model_and_hparams}

\subsection{Model and training details}

In addition to the description in the main paper, we use the following model architectures and training methods:
\begin{itemize}[leftmargin=0.3cm]
    \item All the models are implemented with PyTorch~\citep{paszke2019pytorch}, PyTorch Geometric~\citep{Fey2019Fast}, PyGCL~\citep{Zhu2021tu} and PyTorch Lightning~\citep{Falcon_PyTorch_Lightning_2019}.
    \item When using a bidirectional encoder, half of the hidden feature of 64 is divided and used for each direction. That is, we use 32 for forward edges and 32 for reverse edges.
    \item For positional encoding, we follow the Transformer's original formula~\citep{vaswani2017attention} and set the maximum length of 20. Note that the number of observed nodes is 8. When the numbers of observed nodes are 16, 32, and 64, the maximum lengths are 36, 68, and 132, respectively.
    \item A fixed number ($N^{\obs}$) of observed nodes is sampled at each iteration of the training stage. To add more randomness, we sample a random element from $\{N^{\obs}-2, N^{\obs}-1, N^{\obs}, N^{\obs}+1, N^{\obs}+2\}$ first and select the observed nodes of that number.
    \item For the $\khop$ subgraph, we dropout these edges with the probability of $p_d$~\citep{rong2020dropedge}.
    \item The batch sizes are 16 for \FNTNb and 64 for others, using the gradient accumulation (16 for \FNTN, 1 for \EMUser, and 16 for \HPOMetab).
    \item All model parameters are trained with 16-bit precision supported by PyTorch Lightning~\citep{Falcon_PyTorch_Lightning_2019}.
    \item Each experiment is done on a single GPU. These GPUs are the GeForce GTX 1080Ti, GeForce RTX 2080Ti, and Quadro RTX 8000, but each experiment does not require a specific GPU type. One machine has a total of 40 -- 48 cores of CPUs and 4 -- 8 GPUs.
\end{itemize}

\subsection{Hyperparameter selection}

We tune a subset of hyperparameters with validation sets using Optuna~\citep{akiba2019optuna}. We choose different tuning algorithms and subsets of hyperparameters depending on models and experiment conditions. For baselines (non-InfoMax models), only weight decay is tuned. For SubGNN, we compare the model with all (neighborhood, structure, and position) channels and models with only one channel each. For each case, we tune weight decay, an aggregator for initializing component embedding, $k$ for $\khop$ neighborhood of subgraph component, and numbers of structure anchor patches, border/internal position anchor patches, border/internal neighborhood anchor patches, and LSTM layers for structure anchor patch embedding. Lastly, we tune weight decay, $\lambda$s in MI losses, the ratio in $\topk$ pooling, and DropEdge probability $p_d$ of $\khop$ subgraph for our models.

We use the Tree-structured Parzen Estimator algorithm under a total budget of 50 runs for most experiments. For SubGNN, we choose the random search following the original implementation. Exceptionally, we perform a grid search to evaluate the performance by the number of observed nodes (Figures~\ref{fig:num_obs_xx} and \ref{fig:num_obs_eval}). We run a total of 36 experiments on the space of three $\lambda$ ($\{ 1.0, 2.0, 3.0 \}$), three $\lambdakhop$ ($\{ 1.0, 2.0, 3.0 \}$), two weight decay ($\{ 10^{-4}, 10^{-3} \}$ for \FNTNb and $\{ 10^{-6}, 10^{-5} \}$ for \EMUser), and two pool ratio ($\{ 10^{-3}, 10^{-2} \}$). All hyperparameters are reported in the code.

\section{Discussion on Performance by the Number of Observed Nodes}\label{appendix:discussion_observed_nodes}

In Figure~\ref{fig:num_obs_xx}, we show the performance of $k$-hop PSI + PS-InfoGraph depending on the number of observed nodes. By observing more nodes, the performance on \EMUserb increases but that on \FNTNb decreases. We claim that the impact of performance degradation from neighborhood noise is more significant than information gain from additional nodes in \FNTN.

Where the boundary of the full subgraph is unknown, the increase in the number of observed nodes presents two challenges—first, the number of nodes in the sampled $k$-hop neighborhood increases. Second, the number of nodes in the subgraph but unknown yet to the model decreases. We expect that the performance increases when the information gained from the additional nodes exceeds the noises from the above challenges.

Next, we show that initial nodes are relatively important for \FNTNb dataset. In Table~\ref{tab:result_by_obs_appendix}, we report the experimental result of the GraphSAGE model by the ratio of observed nodes. We call this the $x$\% setting similar to the 100\% setting but uses only $x$\% of nodes in training and evaluation. We set $x$ to 12.5, 25, and 100. As the number of observed nodes decreases, the performance of the GraphSAGE model for all datasets generally decreases. However, the degree varies by dataset. Compared to \EMUserb and \HPOMetab, additional observed nodes in \FNTNb do not significantly affect representation quality. This is in line with \citet{bian2020rumor}.

We demonstrate that the information gained from additional nodes in \FNTNb is relatively small. Considering the challenges from these additions, this explains why the performance on \FNTNb decreases as the number of observed nodes increases in Figure~\ref{fig:num_obs_xx}.

\begin{table}[]
\centering
\caption{Summary of accuracy (5 runs) of GraphSAGE model on three datasets with regard to the ratio of observed nodes at the training and test (i.e., $x$\% setting).}
\label{tab:result_by_obs_appendix}
\vspace{-0.2cm}
\begin{tabular}{lllll}
\hline
\begin{tabular}[c]{@{}l@{}}The ratio of\\ observed nodes\end{tabular} & \FNTN      & \EMUser     & \HPOMetab  \\ \hline
12.5\%                      & 85.9 $\pm$ 1.3 & 54.5 $\pm$ 19.4 & 34.2 $\pm$ 2.1 \\
25\%                        & 86.3 $\pm$ 0.7 & 82.6 $\pm$ 3.5  & 41.2 $\pm$ 1.3 \\
100\%                       & 86.3 $\pm$ 0.7 & 82.1 $\pm$ 1.2  & 47.7 $\pm$ 3.3 \\ \hline
\end{tabular}
\vspace{-0.1cm}
\end{table}

\begin{table}[]
\centering
\caption{Mean wall-clock time (seconds) per batch of the training process on real-world datasets.}
\label{tab:time}
\vspace{-0.2cm}
\begin{tabular}{llll}
\hline
Model           & \FNTN & \EMUser & \HPOMetab \\ \hline
MLP             & 0.021 & 0.040   & 0.018     \\
GraphSAGE       & 0.028 & 0.037   & 0.019     \\
SubGNN          & N/A   & 0.126   & 0.086     \\ \hline
$k$-hop PSI     & 0.816 & 0.103   & 0.406     \\
PS-InfoGraph    & 0.047 & 0.053   & 0.033     \\
$k$-hop PSI + PS-InfoGraph & 0.834 & 0.141   & 0.413     \\ \hline
\end{tabular}
\vspace{-0.1cm}
\end{table}

\section{Training Time}

In Table~\ref{tab:time}, we report the mean wall-clock time per batch of $k$-hop PSI, PS-InfoGraph, and their two-stage models using a single machine (40-core CPU with one GTX1080Ti GPU). For all experiments, we use a batch size of four.

The PS-InfoGraph does not differ much from the baseline MLP and GraphSAGE in training time. The overhead is below 0.03s for all datasets. However, the model using $k$-hop PSI shows a relatively large training time compared to others ($\times 3$ -- $ \times 30$). We confirm that most of the increments occur from $k$-hop sampling. The more edges (e.g., \FNTN) or density (e.g., \HPOMetab) of the global graph, the more time it takes to run. The \EMUserb dataset with a relatively low value for these properties takes a similar training time to that of SubGNN.

\section{Ethical Considerations}\label{appendix:ethical_considerations}

Learning subgraphs requires collecting more attributes (i.e., a global graph plus subgraphs) than learning nodes, edges, and graphs. This could lead to privacy invasion depending on the use case. For example, if we set the global graph as a user network of a social media like \FNTNb and \EMUser, our model should follow up the entire network throughout its life cycle of training and evaluation.

Furthermore, a deeper understanding of the partial subgraph learning problem may enable harmful applications, such as tracking users on social media. Indeed, our study deals with the profiling task of users' gender (\EMUser). Similar concerns are raised in the SubGNN paper, which proposed the original dataset (See Broader Impact section in \citet{alsentzer2020subgraph}). Also, while our model suggests the positive application of fake news detection, it leaves room for attacks to deceive. This is a general problem with any machine learning model, and thus researchers accessing and using this research must be mindful of potential harm.

Lastly, \EMUserb dataset simulates the prediction task of binary genders (male and female), but genders could be non-binary in reality. Future research should consider that \EMUserb is an over-simplified dataset for benchmarking purposes.

\clearpage

\fi

\end{document}